\documentclass[letterpaper, 10 pt, conference]{ieeeconf}

\IEEEoverridecommandlockouts
\usepackage{amsmath, amsfonts, amssymb} 
\usepackage{array}
\usepackage{graphicx}
\usepackage{xspace}
\usepackage{stmaryrd}
\usepackage{xcolor}
\usepackage{mathtools}
\usepackage{float}
\usepackage{textcomp}
\usepackage{multirow}
\usepackage{diagbox}
\usepackage[printonlyused,withpage,nolist,nohyperlinks]{acronym}
\usepackage{subcaption}
\usepackage{caption}
\usepackage{lipsum}
\usepackage{algorithm}
\usepackage{algpseudocode}  
\usepackage{booktabs}
\usepackage{comment}
\usepackage{stfloats}
\usepackage{placeins}
\usepackage{url}
\usepackage{microtype}
\usepackage{makecell}
\usepackage{amsmath}
\usepackage{url}

\begin{document}
\title{A Collaborative Reasoning Framework for Anomaly Diagnostics in Underwater Robotics}

\author{Markus Buchholz$^{1}$, 
         Niamh Ellis$^{1}$,
         Rahaf Abu Hara$^{1}$,
         Ignacio Carlucho$^{1}$, and
         Yvan R. Petillot$^{1}$ %
\thanks{$^{1}$School of Engineering \& Physical Sciences, Heriot-Watt University, Edinburgh, UK
        {\tt\small nmfe2000@hw.ac.uk}}%
}
  
\maketitle
\begin{abstract}
The safe deployment of autonomous systems in safety-critical environments requires a paradigm that synergises human expertise with AI-driven data analysis, particularly for unforeseen anomalies. This paper introduces AURA (Autonomous Resilience Agent), a collaborative framework for anomaly and fault diagnostics in robotics.
AURA integrates Large Language Models (LLMs), a Digital Twin (DT), and human-in-the-loop interaction to identify and respond to anomalous behaviour in real time.
The architecture employs a two-agent design with a distinct separation of responsibilities: i) a low-level \emph{State Anomaly Characterisation Agent} that monitors telemetry and translates data into a structured natural-language problem description, and ii) a high-level \emph{Diagnostic Reasoning Agent}, which engages a human operator in an interactive, knowledge-grounded dialogue to determine the root cause, supported by external knowledge sources.
Crucially, the final, human-validated diagnosis is processed into a new reference example that is used to inform the low-level agent when it encounters a similar anomaly. 
This feedback loop allows the operator's expert knowledge to be available to 
the AI, transforming it from a static tool into an adaptive partner. We detail the framework's operational principles and a concrete implementation, establishing a new pattern for creating trustworthy, continuously improving human-robot teams.\footnote{%
AURA project webpage: \url{https://markusbuchholz.github.io/aura.html}}
\end{abstract}


\section{INTRODUCTION}

Remotely Operated Vehicles (ROVs) operate in unpredictable environments where unforeseen anomalies can compromise mission success. Traditional fault-tolerant control techniques and model-based Fault Detection \& Identification (FDI) provide robustness for anticipated failures, but they struggle with novel or compound faults beyond their design assumptions. The advent of advanced AI, particularly Large Language Models (LLMs), offers a new paradigm for anomaly reasoning by leveraging vast knowledge and cognitive reasoning ability \cite{cui2024fault}. However, deploying LLMs directly as autonomous decision-makers in safety-critical robotics raises serious trust, reliability, and transparency concerns. Recent studies highlight the unpredictable behaviour and failure modes of LLM-based systems, including outages and erratic outputs in critical applications \cite{chu2025outages}. Robots controlled naively by LLMs or Vision-Language Models (VLMs) have shown vulnerabilities such as brittleness to distribution shifts and susceptibility to adversarial inputs \cite{wu2024vulnerability}. These issues highlight that fully replacing human operators with \emph{black-box} LLM intelligence is still unattainable for safety-critical domains.
However, rather than seeking to eliminate humans, the emerging consensus is to embed human expertise in the AI loop to ensure reliability.

\begin{figure}[t]
  \centering
  \includegraphics[width=0.95\columnwidth]{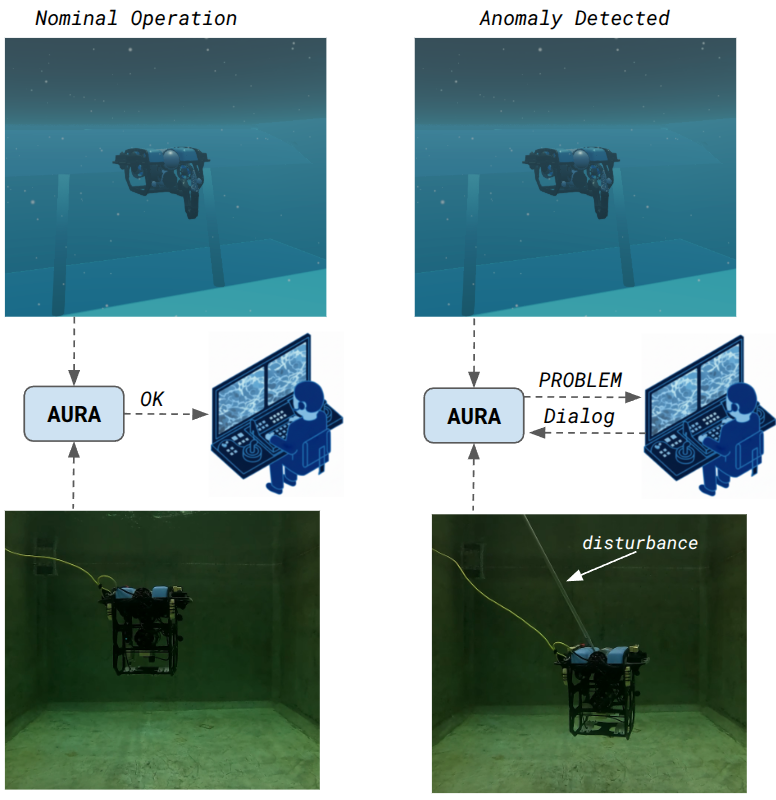}
\caption{AURA uses a DT  as a real-time baseline for the physical ROV. When a difference occurs between the two, an anomaly detection is triggered, which initiates a collaborative dialogue to diagnose the problem. These experiences are used in the future to enhance the AI agents.}
  \label{fig:aura_overview}
  \vspace{-0.5cm}
\end{figure}

This paper introduces AURA (Autonomous Resilience Agent), a collaborative anomaly diagnostic framework between a human operator and an AI system. 
AURA follows the design philosophy of the Underwater Robot Self-Organizing Autonomy (UROSA) framework \cite{buchholz2025urosa}, which establishes a distributed ecosystem of specialised, agentic AI nodes for underwater robotics.
%
%
AURA utilises a Digital Twin (DT) alongside LLM-based agents to detect anomalies and propose potential fault hypotheses in real time. We illustrated the high-level concept of our architecture in Fig. \ref{fig:aura_overview}. When an anomaly occurs, the behaviour of the physical ROV diverges from its DT, triggering a collaborative diagnostic process.
AURA employs a two-agent architecture with a clear division of roles, as shown in Fig. \ref{fig:aura_architecture}. A low-level \emph{State Anomaly Characterisation Agent} (Agent A) monitors the ROV’s telemetry against its digital twin to detect statistical deviations, then translates complex numerical anomalies into a structured natural-language description of the problem state. A high-level \emph{Diagnostic Reasoning Agent} (Agent B) then engages the human operator in an interactive dialogue to diagnose root causes, supported by relevant external knowledge sources. Crucially, the final human-validated diagnosis is stored in the system as a labelled example. Through this loop, the human operator’s expertise is progressively distilled into Agent A’s perceptual model. In essence, each anomaly encounter makes the framework better prepared for the next, establishing a pathway for trustworthy autonomy that directly incorporates human feedback.

The main contributions of this work are:
\begin{itemize}
    \item A human-AI collaborative framework for anomaly diagnostics that ensures operator authority while utilising the vast knowledge and cognitive reasoning of LLMs in determining root causes of encountered anomalies
    \item A two-agent architecture which separates anomaly detection and root cause diagnosis, improving transparency and controllability
    \item A retrieval-augmented diagnostic framework which accumulates validated anomaly-solution pairs to support experience-based reasoning and progressive improvement in future anomaly diagnostic scenarios 
    \item A natural language method of anomaly diagnostics creating a user-friendly interface and making it easier for non-robotic experts to diagnose anomalies when using ROVs  
\end{itemize}
We present results on a real-time robotic platform that showcase the capability of our method to detect, diagnose, and improve its ability over time as the system interacts with the human operator.

\section{LITERATURE REVIEW}


Historically, ensuring reliability in autonomous robots relied on deterministic models. Classical fault diagnosis and Fault-Tolerant Control (FTC) in Autonomous Underwater Vehicles (AUVs) used analytical redundancy and state observers to detect failures \cite{fossen2011handbook}. These methods are effective for anticipated failure modes but are brittle when confronting anomalies that fall outside predefined model assumptions. Digital twins improve anomaly detection by simulating expected nominal behaviour in parallel with the real AUV, enabling real-time detection of deviations \cite{isaku2025digital}. Recent studies have leveraged digital twins for fault diagnosis in marine systems and robot joints, often using neural networks to flag discrepancies as potential faults \cite{song2023digital}. However, once an anomaly is detected, classical methods lack the cognitive reasoning to hypothesise about unmodeled failure causes.

LLMs have emerged as a pathway to endow robots with flexible reasoning \cite{bubeck2023sparks}. Researchers have begun exploring LLMs as decision-makers for robotics, with promising early results \cite{vemprala2023chatgpt}. Google’s SayCan framework grounded an LLM’s text outputs in executable policies \cite{ahn2022do}, while PaLM-E demonstrated an embodied multimodal language model that integrates vision and language for robotic control \cite{driess2023palme}. A landmark study by Mon-Williams et al. (2025) introduced an embodied LLM-based system (ELLMER) that uses GPT-4 to perform long-horizon tasks \cite{mon-williams2025ellmer}. These works illustrate the potential of LLMs as \textit{brains} for robots. At the same time, deploying such models highlights significant challenges: unpredictability, lack of grounding, and the tendency of LLMs to hallucinate or act unsafely if not properly constrained \cite{wu2024vulnerability}. The consensus in recent literature is that LLMs can greatly augment robotic intelligence if their reasoning can be grounded, verified, and kept in check by human oversight.

In designing AURA, we draw inspiration from cognitive architectures like SOAR that emphasize a separation of perception and high-level reasoning \cite{laird2012soar}. This principle prevents high-dimensional noisy sensor data from overwhelming cognitive processes. Recent robotics research echoes this: many LLM-based frameworks isolate the language model’s role as a planner, while delegating real-time control to traditional controllers. In our context, a clear division is beneficial: Agent A (perception) operates on raw telemetry, while Agent B (reasoning) operates on symbolic information to hypothesise causes. By having Agent A convert sensor data into an intermediate, human-interpretable form (a \textit{problem characterisation}), we achieve a form of symbolic grounding, allowing Agent B to deal with a clean, distilled representation of the problem.

Human-in-the-loop learning has gained traction as a means to align AI systems with human expertise. The success of Reinforcement Learning from Human Feedback (RLHF) demonstrates the power of using human preference data to steer AI behaviour \cite{ouyang2022training}. This process, however, is computationally intensive and not suitable for on-the-fly learning during a mission. Recent work by Liu et al. (2022) introduced Sirius, a framework where a robot’s policy is continually updated from human interventions during execution \cite{liu2022sirius}. AURA’s Human-in-the-Loop Distillation follows a similar ethos but in the context of fault diagnosis. Each time the human operator guides Agent B to a correct diagnosis, that experience is stored in the VDB to enhance Agent A’s mapping from raw signals to problem descriptions. Our method is one experience at a time, fitting into the concept of online or lifelong learning in robotics. We address the challenge of continual updates by storing distilled examples in a vector database and employing them via retrieval-augmented generation (RAG) \cite{lewis2020rag}, rather than retraining Agent A and hence avoiding a higher computational cost. This strategy allows Agent A to effectively access past human-confirmed cases when a new anomaly is encountered, guiding its characterization.


\section{THE AURA COLLABORATIVE REASONING ARCHITECTURE}

\begin{figure}[t]
  \centering
  \includegraphics[width=0.95\columnwidth]{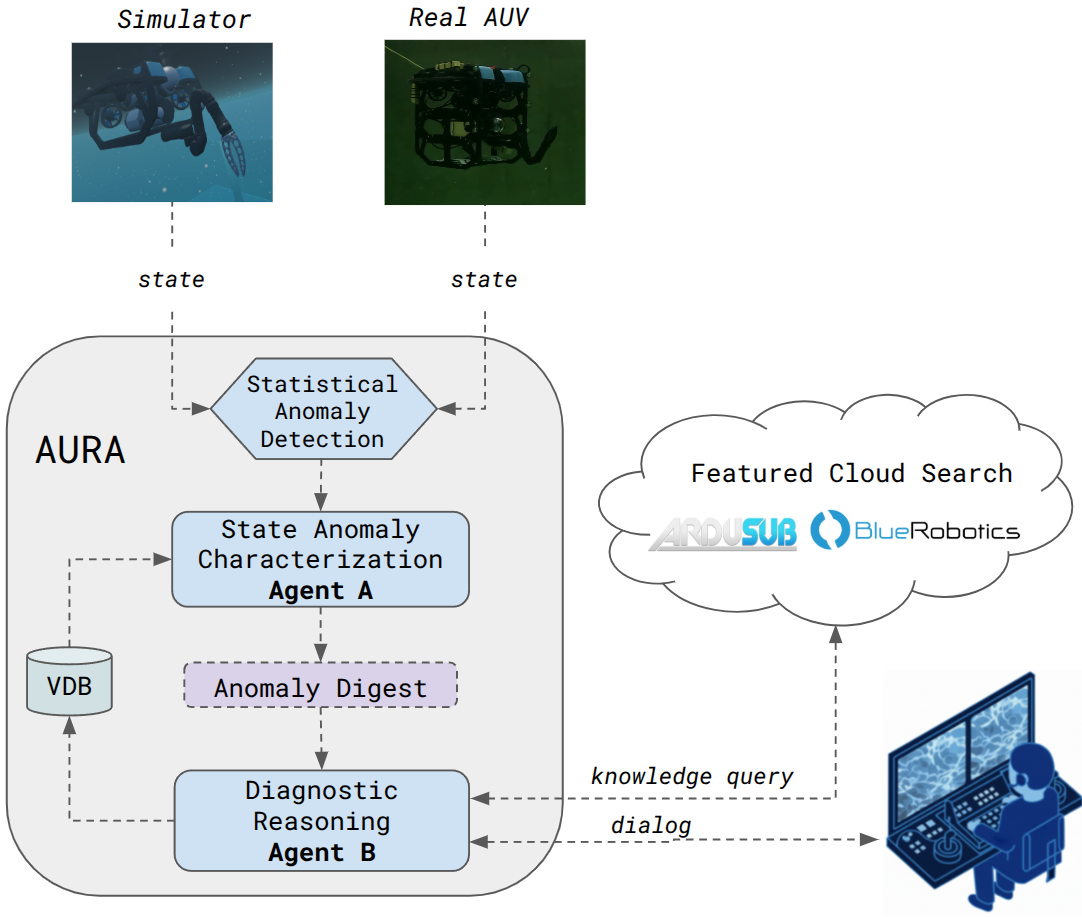}
  \caption{The AURA Collaborative Reasoning Architecture. Anomaly detection triggers the State Anomaly Characterisation Agent (A), which translates raw data into a structured Problem Characterisation. This is passed to the Diagnostic Reasoning Agent (B), which uses external knowledge and an interactive dialogue with the operator to find a solution. The outcome is stored in the VDB to refine Agent A.}
  \label{fig:aura_architecture}
  \vspace{-0.4cm}
\end{figure}

AURA is built as a two-agent system operating in tandem with a human operator. The architecture (Fig. \ref{fig:aura_architecture}) is organised into sequential stages, from anomaly detection to human-guided diagnosis and learning feedback. The design philosophy is to keep each agent’s role interpretable and constrained, while enabling rich interaction with the human and external knowledge sources. In the following subsections, we provide a general overview of each of the components of AURA.

\subsection{Digital Twin Architecture}
The foundation of AURA's diagnostic capability is a digital twin of the vehicle, which serves as a dynamic, real-time \emph{normative model}. For each operation, this digital twin runs in parallel to the physical vehicle, receiving the exact same command inputs. The AURA framework continuously monitors and compares the state telemetry from both the physical asset and its virtual replica via ROS 2 topics. This parallel monitoring provides a constant baseline of expected behaviour, allowing the system to detect deviations that would otherwise be lost in sensor noise.

The digital twin is a physics-based model developed in the Stonefish simulator \cite{s2}, which is fully integrated into the vehicle's ROS 2 computational pipeline and interfaces. Stonefish models the coupled 6-DOF rigid-body dynamics of the vehicle, including geometry-based approximations of hydrodynamic added mass, damping, and buoyancy forces, together with thruster and sensor models \cite{s2}. It has previously been found to be a strong contender for use in digital twin applications \cite{UnderwaterSimsForDT}, has compared favourably against other open-source underwater simulators in independent benchmarking \cite{huang2024urobench}, and remains one of the most advanced and actively maintained open-source underwater simulators \cite{s2}.

It is important to note that AURA does not require the digital twin to be a perfect replica of the physical vehicle. Like any physics-based simulator, Stonefish relies on approximated hydrodynamic coefficients, and a systematic sim-to-real gap is therefore unavoidable. Our framework is explicitly designed to be robust to this: anomalies are not flagged from raw twin-vehicle differences, but from \emph{statistical} deviations of the residuals relative to their empirically characterised nominal distribution (Section III-B). Constant or slowly varying model mismatch is thus absorbed into the nominal residual statistics, and only abrupt, fault-induced departures from this baseline trigger the diagnostic pipeline. In this sense, Stonefish provides a normative reference for vehicle-level behaviour in bounded operational scenarios, supporting event-level anomaly detection rather than long-duration environmental prediction.

AURA is implemented as a software suite that interfaces with the ROV’s ArduSub autopilot, leveraging its MAVLink telemetry streams. 
AURA’s anomaly detection node subscribes to the vehicle’s state estimate and a parallel digital twin.

\subsection{Statistical Anomaly detection}

By comparing key state vectors from the real vehicle and the twin in real-time, the node computes error residuals. If any residual exceeds adaptive thresholds, an anomaly trigger is issued.
A normative model of expected vehicle behaviour is first established by generating a multivariate statistical distribution from numerous simulation runs of the digital twin. This model captures the mean state vector $\mu$ and the covariance matrix $\Sigma$ of key operational parameters. The live vehicle's state vector, $x$, is continuously compared against this model using the Mahalanobis Distance, calculated as $MD^2 = (x-\mu)^{T} \Sigma^{-1} (x-\mu)$, where $(\cdot)^{T}$ denotes the vector transpose. A statistically significant deviation, where $MD^2$ exceeds a predefined threshold based on the chi-squared ($\chi^2$) distribution, is flagged as an anomalous event.

In open-water deployment, environmental effects (e.g., currents and turbulence) may slowly shift residual statistics. To accommodate this non-stationarity, AURA adaptively updates the nominal residual mean and covariance using exponentially weighted estimation. Let $r_t = x_{\text{real},t} - x_{\text{twin},t}$ denote the residual at time $t$. During operator-verified nominal operation, updates are performed as
\begin{equation}
\mu_t = (1 - \alpha)\mu_{t-1} + \alpha r_t,
\end{equation}
\begin{equation}
\Sigma_t = (1 - \alpha)\Sigma_{t-1} + \alpha (r_t - \mu_t)(r_t - \mu_t)^{T},
\end{equation}
where $\alpha \in (0,1)$ is the adaptation rate. Residuals from confirmed fault events are excluded. This preserves sensitivity to abrupt fault-induced covariance changes while mitigating false positives from gradual, structurally smooth environmental drift. 

\subsection{Agent Stacks and Rationale}

Given the complexity of translating raw telemetry into meaningful diagnostic insight, AURA adopts a dual-LLM architecture consisting of a State Anomaly Characterisation Agent (Agent A) and a Diagnostic Reasoning Agent (Agent B). Rather than relying on a single large model, responsibilities are divided to balance computational efficiency and cognitive load: high-frequency signal-to-symbol translation is handled by a lightweight local model (4–12B parameters), while the more computationally intensive 27B model is reserved for low-frequency, high-level causal reasoning once an anomaly is triggered. This separation enables scalable deployment on edge hardware without compromising reasoning capability.



The \textbf{State Anomaly Characterisation Agent} (Agent A) is realised using a moderate-sized language model (in the 4-12 billion parameter range) served locally via an open-source inference server in this case Ollama. This choice ensures that raw vehicle telemetry never leaves the local network. 

\begin{figure*}[tp]
    \centering
    \includegraphics[width=0.95\textwidth]{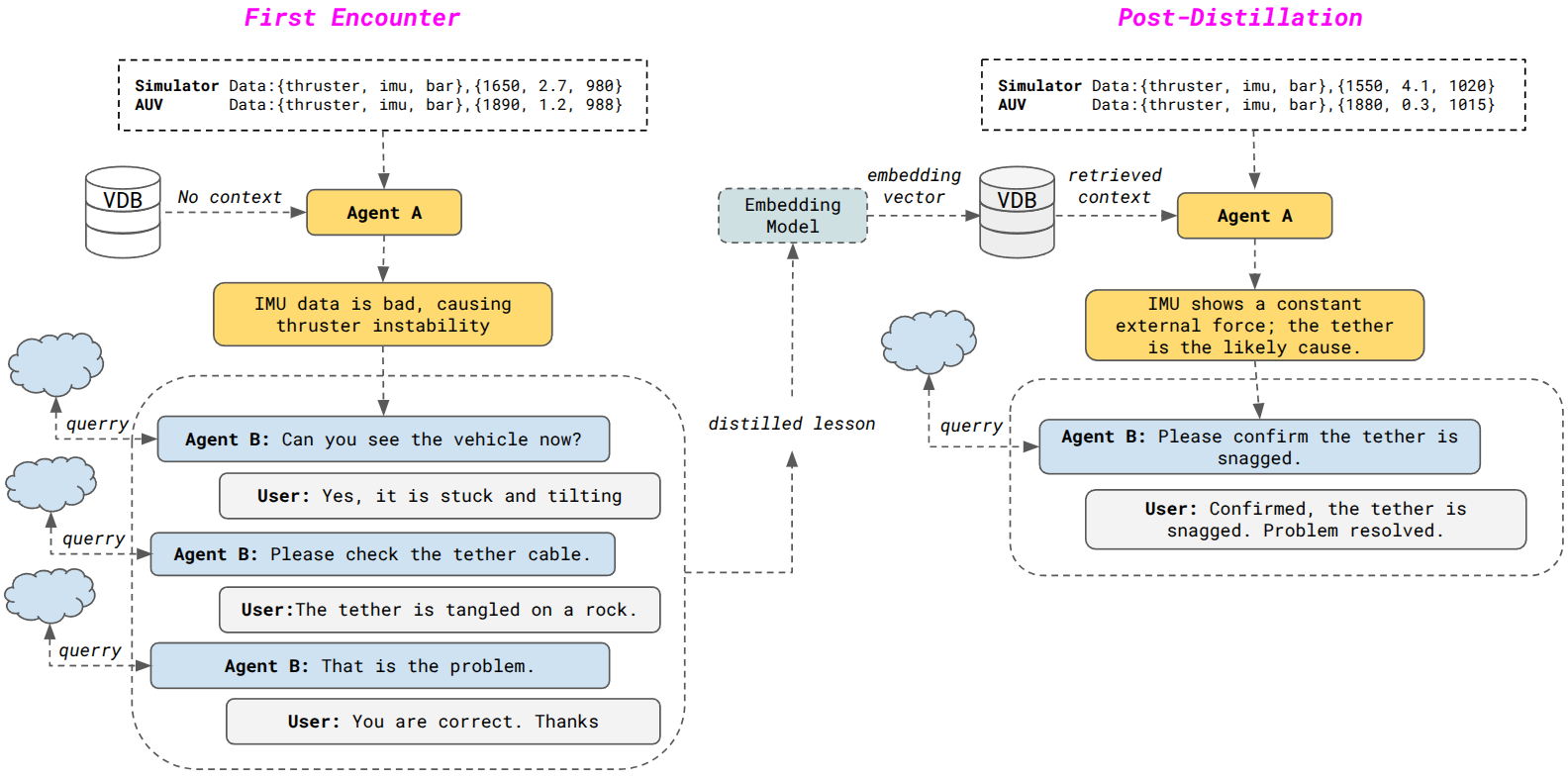}
    \caption{An illustrative example of the Human-in-the-Loop Distillation process, designed to produce a quantifiable improvement in diagnostic performance. The figure contrasts the workflow in two phases: \textbf{First Encounter} (Left) and \textbf{Post-Distillation} (Right). In the First Encounter, with \textit{No context} from the VDB, Agent A produces a generic characterization. This necessitates an extended, multi-turn diagnostic dialog between Agent B and the operator to identify the root cause. The outcome of this session is processed into a \textbf{distilled lesson}, which is then passed to an \textbf{Embedding Model}. The resulting \textbf{embedding vector}, a numerical representation of the experience, is stored in the VDB. In the Post-Distillation phase, when a similar anomaly occurs, Agent A receives \textbf{retrieved context}. This allows it to generate a highly specific and insightful characterization, leading to a much shorter, confirmatory dialog and demonstrating the system's enhanced diagnostic efficiency.}
    \label{fig:distillation_example}
\end{figure*}

The agent's behaviour, illustrated in Fig. \ref{fig:distillation_example}, is powered by the knowledge and memory systems. 
The Vector Database (VDB) is implemented using a lightweight local vector store (ChromaDB). Textual \textit{distilled lessons} are converted into numerical embeddings using a locally hosted sentence-transformer model (\texttt{nomic-embed-text}), enabling similarity-based retrieval across semantically related but non-identical anomaly descriptions. The entire lifecycle of the system's experiential memory, from creation to retrieval, is handled offline, creating a fully private, field-deployable learning loop.


The \textbf{Diagnostic Reasoning Agent} (Agent B) requires more advanced cognitive capabilities and hence this role is filled by a larger, more powerful open-source model (Gemma 3 27B), also served locally via Ollama. The agent's conversational flow and its ability to use tools are managed by an orchestration framework.
This agent is granted a single, highly constrained tool: access to the \emph{Featured Cloud Search} function for knowledge grounding.

The implementation of the AURA agents prioritises data privacy, low latency, and operational independence from cloud services-all critical requirements for deployment in remote or secure environments. To meet these principles, our stack is centred on locally-hosted, open-source models and libraries.

\subsection{Human-in-the-Loop Interface}
The operator's interactions with AURA are managed directly through ROS 2 topics, ensuring seamless integration into the existing robotics ecosystem. The nature of the interaction evolves as the system learns. As illustrated in Fig. \ref{fig:distillation_example}, initial encounters with novel faults typically involve a longer, exploratory dialogue where the operator provides critical context. Subsequent encounters with similar faults often require only a brief, confirmatory exchange, as Agent A's improved characterisation allows Agent B to present the most credible root cause based on previous instances. The design presents these AI-generated insights as a cognitive support tool that operates in parallel with the primary piloting and telemetry displays to minimize cognitive load on the operator.
Additionally, the human in the loop functions as the authority within our framework and hence false alarms, where the system is triggered, but there is no anomaly, will be resolved via human intervention. 

\subsection{Architectural Safeguards and Verifiability}
The AURA framework ensures safety and reliability by having Agent B act as an advisory component and being incapable of executing physical actions. In our collaborative method, the human operator has oversight and is in charge of any action taken on the vehicle in response to anomalies. The human operator is the final arbiter of any diagnosis. This human validation is critical for the integrity of the system's long-term memory. The process shown in Fig. \ref{fig:distillation_example} is a direct application of this principle: the lengthy diagnostic session on the left is filtered and validated by the operator to produce a single, high-quality training example for the VDB. This \emph{Human-in-the-Loop Distillation} process acts as a verifiability filter, preventing the propagation of flawed reasoning into the system's experiential memory and ensuring the AI utilises information only from verified, expert-validated examples. 

\section{The AURA Collaboration Workflow}

In this section, we provide an overview of the AURA's workflow, which we divide into stages. 


\subsection{Stage 1: Anomaly Detection and Signal Interpretation}
The process begins when the Digital Twin's expected behaviour differs from that of the real vehicle. These events trigger a quantitative anomaly detection step, and in turn 
%
activates the \textbf{State Anomaly Characterisation Agent} (Agent A). 
Agent A's primary task is signal-to-symbol translation, converting high-dimensional sensor disparities coming from the DT and the real-vehicle into semantically meaningful problem characterisations, enabling flexible similarity matching across partially overlapping anomaly patterns beyond fixed, predefined feature mappings. The agent is implemented as a specialised LLM, grounded solely in vehicle data and past validated experiences.

Initially, with an empty Vector Database (VDB), Agent A functions as a basic data transcriber. It takes the raw numerical anomaly signature and produces a factual but uncritical description of the observable symptoms. Specifically, the input prompt to Agent~A contains a structured JSON array encoding the deviation between the Digital Twin and the real vehicle. This representation includes key telemetry variables such as \texttt{SERVO\_OUTPUT\_RAW} for all thrusters, \texttt{VFR\_HUD} (e.g., heading, airspeed, groundspeed), and \texttt{GLOBAL\_POSITION\_INT} (e.g., relative altitude), thereby providing a grounded, machine-readable summary of the state discrepancy. For instance, given a heading error, its initial \emph{Problem Characterisation} might be a simple statement:
\textit{Observed vehicle heading is 35 degrees, while expected heading is 2 degrees.}
This initial output is useful but lacks depth. The real power of the AURA framework is realised after the VDB is populated through the \emph{Human-in-the-Loop Distillation} process. Once a similar fault has been diagnosed in a previous session, where the human operator 
confirmed the cause was magnetic interference, a high-quality training example is stored in the VDB.
On subsequent encounters with a similar anomaly, Agent A retrieves this distilled lesson. Now, its output is no longer a simple transcription but a critical characterisation that reflects the captured human expertise:
\textit{A critical anomaly was detected in the IMU compass reading. The observed vehicle heading of 35 degrees is a massive deviation from the expected 2 degrees. This pattern is consistent with previously validated instances of magnetic interference. Potential root causes include miscalibration or proximity to large metallic structures.}
This enhanced characterisation is far more valuable. By retrieving and reasoning from the distilled knowledge, Agent A now provides crucial context that primes Agent B and the human operator for a much faster and more effective diagnostic process. This clearly demonstrates how the collaborative dialogue and external knowledge accessed in Stage 2 are instrumental in building a more intelligent perception layer in Stage 1.

\subsection{Stage 2: Knowledge-Grounded Diagnostic Reasoning}
The \textit{Problem Characterisation} from Agent A serves as the symbolic input to the \textbf{Diagnostic Reasoning Agent} (Agent B), the primary cognitive engine of the framework. The agent's core task is to formulate and evaluate causal hypotheses based on this structured description of the system's anomalous state.
Upon receiving the characterization, Agent B immediately performs a critical knowledge grounding step. It programmatically formulates and executes queries against a curated set of external, user-chosen, high-authority data sources, such as technical manuals and engineering forums. This retrieval-augmented mechanism is a fundamental design choice to mitigate model hallucination and anchor the reasoning process in verifiable, domain-specific, and up-to-date information.
After synthesising the problem characterisation with the retrieved external knowledge, Agent B generates a ranked set of plausible hypotheses. The highest-probability hypothesis, along with its evidential support, is presented to the human operator. The system then enters an interactive refinement loop. Through the dialog interface, the operator can provide additional contextual data unavailable to the sensors, challenge the agent's premises, or direct it to explore alternative lines of reasoning. 
This iterative, human-guided process continues until a diagnostic conclusion is collaboratively reached and validated by the operator, thereby synergising the AI's rapid data synthesis with the operator's contextual expertise. 


\subsection{Stage 3: Human-in-the-Loop Distillation and Learning Feedback}
This stage distinguishes AURA as a continually improving system. After the diagnostic session concludes, the entire sequence of events is logged and processed into a new training example, including the raw telemetry, Agent A’s description, the conversation transcript, and the final confirmed diagnosis. From this log, we extract a concise training tuple for Agent A: (raw anomaly data $\rightarrow$ correct problem characterisation). This tuple is inserted into the VDB as a high-quality, operator-validated reference example. Only confirmed distilled tuples are stored, reducing the risk of propagating incorrect intermediate reasoning. When a similar anomaly is encountered, Agent A queries the VDB and retrieves semantically aligned prior cases to inform its output. Because retrieval is based on similarity ranking across all stored cases rather than deterministic rule matching, no single erroneous entry can systematically dominate future characterisations. This experience-retrieval mechanism echoes memory-augmented approaches recently demonstrated for underwater manipulation \cite{buchholz2025context}.


\subsection{Stage 4: Proactive Pre-Mission Knowledge Injection}
While the core loop of AURA occurs during missions, we have also developed a mechanism for pre-mission tuning. Before deploying a vehicle, an operator can engage AURA in a simulated dialogue about known issues. For instance, an operator could present Agent A with log data from a historical mission and walk through the diagnosis with Agent B. The outcome is fed into the VDB just like a live case. This will be investigated further in future work.



\begin{figure*}[tp]
    \centering
      \includegraphics[width=0.65\textwidth]{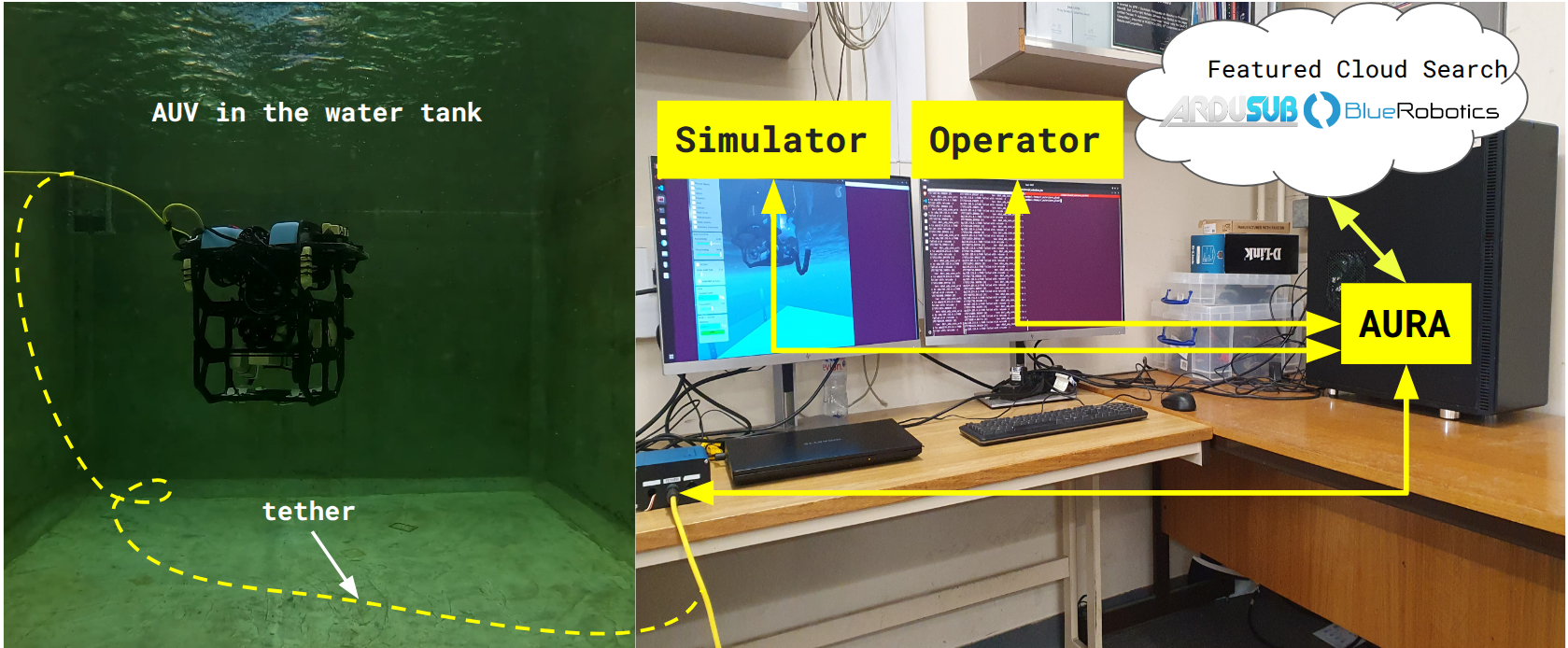}
    \caption{The experimental validation platform for the AURA framework. The system simultaneously processes state telemetry from two sources: a physical BlueROV2 operating in a water tank (the real system) and a digital twin running in the Stonefish simulator (the normative model). Both data streams are fed into the AURA framework, which is managed on a local host station. This dual-reality setup allows for the precise generation and detection of anomalies, providing a controlled environment to evaluate the performance of the human-AI collaborative diagnostic loop.}
   \label{fig:experimental_setup}
\end{figure*}

\section{EVALUATION}


\subsection{Experimental Protocol}
The experiments were conducted using the physical and virtual assets depicted in Fig. \ref{fig:experimental_setup}. A BlueROV2 was operated in a water tank, while a parallel digital twin was run in the Stonefish simulator. We note that the BlueROV2 is a tethered, remotely operated vehicle rather than a fully autonomous AUV; its piloted operation naturally places a human operator in the loop, which aligns directly with the collaborative diagnostic paradigm proposed in this work. The entire system is managed within a ROS 2 computational graph.
The AURA software, including the locally-hosted LLMs and VDB, runs on a ground station PC with dual GPU support.
Our evaluation methodology was divided into two distinct phases:

\textbf{Phase 1: Knowledge Acquisition (VDB Priming).} To simulate a system with a baseline of operational experience, we first populated the Vector Database. Starting with an empty VDB, we conducted five initial diagnostic sessions. The number of priming sessions was empirically set to five. This specific number was chosen to establish a minimal viable baseline, ensuring the VDB contained at least one to two reference experiences for each of the distinct anomaly classes tested (two thruster disturbances, two rotational, and one vertical), demonstrating the framework's few-shot data efficiency. Each session was a full human-in-the-loop dialog that concluded with a validated diagnosis. The outcome of each of these five sessions was distilled into a unique reference example and stored in the VDB. This process resulted in a \textit{primed} VDB containing five distinct, human-validated experiences, representing a moderately experienced system.

\textbf{Phase 2: Performance Evaluation.} This phase was designed to test the core hypothesis. We measured the system's performance on a set of new, previously unseen anomalies; these validation anomalies belonged to the same general classes as the priming examples but varied significantly in their physical parameters to test generalization (e.g., if a training example involved a tether snag causing a 15-degree starboard list, the validation anomaly involved an unmodeled payload shift causing a 25-degree port pitch). Tether-induced disturbances of this kind are a well-documented operational challenge for tethered underwater vehicles \cite{s3}. 
\begin{itemize}
    \item \textbf{First Encounter:} The system was tested against the new anomalies with a clean, empty VDB. This measures the baseline performance without any prior experience.
    \item \textbf{Post-Distillation:} The system was tested against the same new anomalies, but this time it had access to the VDB primed with the five experiences from Phase 1.
\end{itemize}
This protocol allowed us to directly quantify the performance improvement attributable to the distilled knowledge in the VDB. The entire evaluation process (Phase 1 and Phase 2) was repeated five times (n=5) per anomaly class to demonstrate initial empirical consistency and serve as a proof-of-concept. 

\subsection{Use Cases}
Three classes of anomalies were tested.
\begin{itemize}
    \item \textbf{Thruster Control Under Disturbance:} The stabilized ROV was subjected to various external physical forces.
    \item \textbf{Anomalies in Vertical Motion:} The ROV was commanded to change depth while being physically impeded.
    \item \textbf{Anomalies in Rotational Motion:} The ROV was commanded to rotate while being physically impeded.
\end{itemize}

\begin{figure*}[ht]
    \centering
    \includegraphics[width=0.9\textwidth]{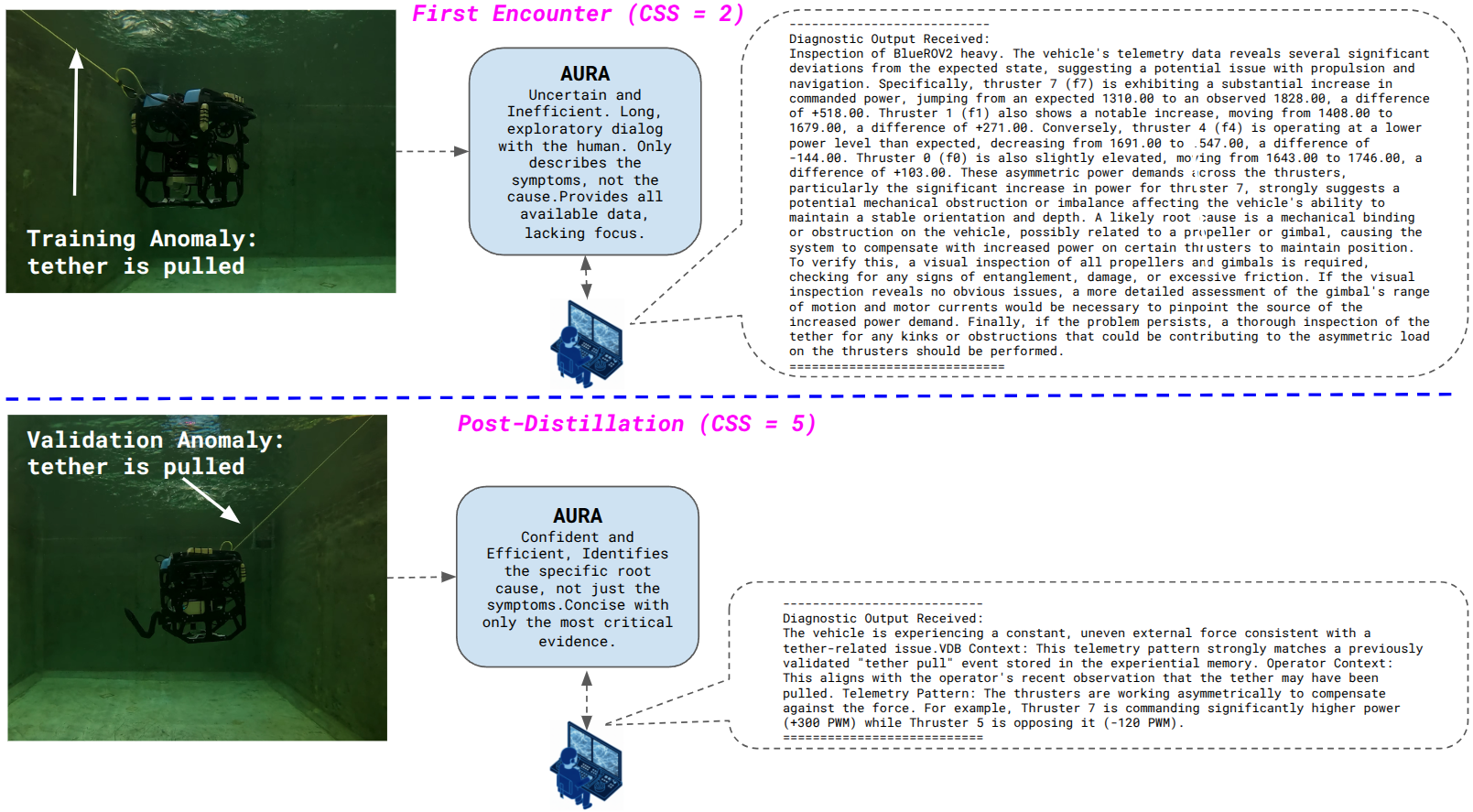}
    \caption{Qualitative comparison of AURA's performance, contrasting its response to a \textbf{Training Anomaly} (Top) with its response to a subsequent \textbf{Validation Anomaly} (Bottom). \textbf{(Top)} In the \textbf{First Encounter}, the system is inefficient and its initial output is a lengthy \textbf{Descriptive Hypothesis (CSS=2)}, which details symptoms but fails to identify the root cause. \textbf{(Bottom)} After the lesson from the first event is distilled, in the \textbf{Post-Distillation} phase, the system is confident and efficient. For a similar validation anomaly, it produces \textbf{Causal Identification (CSS=5)}, correctly pinpointing the tether as the root cause.}
   \label{fig:qualitative_results}
   \vspace{-0.2cm}
\end{figure*}

\begin{table*}[ht]
\caption{Performance Improvement via Human-in-the-Loop Distillation}
\label{tab:results}
\centering
\small
\setlength{\tabcolsep}{2pt}
\renewcommand{\arraystretch}{1.05} 
\begin{tabular}{|p{6cm}|c|c|c|c|}
\hline
\textbf{Use Case} & \multicolumn{2}{c|}{\textbf{First Encounter}} & \multicolumn{2}{c|}{\textbf{Post-Distillation}} \\
\cline{2-5}
\textbf{(Avg. of $n=5$)} & \textbf{CSS (SD)} & \textbf{Turns (SD)} & \textbf{CSS (SD)} & \textbf{Turns (SD)} \\
\hline
Thruster Disturbance & 2.6 ($\pm$0.5) & 6.4 ($\pm$1.3) & 4.6 ($\pm$0.5) & 2.0 ($\pm$0.0) \\
\hline
Vertical Motion Anomaly & 2.9 ($\pm$0.4) & 5.8 ($\pm$1.1) & 4.9 ($\pm$0.3) & 1.6 ($\pm$0.5) \\
\hline
Rotational Motion Anomaly & 2.5 ($\pm$0.5) & 6.5 ($\pm$1.4) & 4.8 ($\pm$0.4) & 1.8 ($\pm$0.4) \\
\hline
\textbf{Overall Average} & \textbf{2.7 ($\pm$0.5)} & \textbf{6.2 ($\pm$1.2)} & \textbf{4.8 ($\pm$0.4)} & \textbf{1.8 ($\pm$0.4)} \\
\hline
\end{tabular}
\end{table*}

\subsection{Results and Discussion}
The performance of AURA was evaluated using two metrics: (i) the number of \emph{Dialog Turns} required to reach a diagnosis, and (ii) the \emph{Characterization Specificity Score (CSS)}. The CSS was assessed on a 5-point Likert scale by a domain expert (5 years of ROV systems experience). 

Due to the small sample size (n=5 per condition), standard deviations are provided alongside average scores to indicate data spread, though these serve as an empirical indicator rather than a strict statistical proof. To reflect a high standard for diagnostic autonomy, the CSS was assigned according to the following predefined rubric:
\begin{itemize}
    \item[1:] \textbf{Incorrect:} The output is factually wrong, misses the primary anomaly, or is too generic to be useful.
    \item[2:] \textbf{Descriptive Hypothesis:} The output accurately describes the  symptoms with quantitative data and suggests a plausible class of causes with  troubleshooting steps.
    \item[3:] \textbf{Targeted Hypothesis:} The output uses historical context from the VDB to identify the single most likely, specific root cause based on past, similar events.
    \item[4:] \textbf{Verifiable Hypothesis:} The output proposes the single most likely cause and suggests a specific test the operator can perform to confirm or deny the hypothesis.
    \item[5:] \textbf{Causal Identification:} The output can, from telemetry alone, identify the exact root cause with extremely high confidence, requiring only final confirmation from the operator.
\end{itemize}
As this study represents an initial proof-of-concept, evaluations were conducted by a single expert, precluding formal inter-rater reliability analysis. Validating CSS objectivity across independent operators will be addressed in future field trials.
%
The diagnostic dialog was terminated once a mutually agreed-upon hypothesis was reached with over 90\% operator confidence. The results, summarized in Table \ref{tab:results}, show a significant improvement in performance when the system had access to the primed VDB.

During the \textbf{First Encounter} (with an empty VDB), Agent A produced characterizations that were factually correct but generic, achieving an average CSS of only 2.7. As shown in the example in Fig. \ref{fig:qualitative_results} (left panel), this low-specificity output typically scored a 2 on our rubric, requiring a lengthy dialog (averaging 6.2 turns) for the operator to guide Agent B to the root cause.
The performance in the \emph{Post-Distillation} condition was dramatically different. With access to the VDB primed with varied experiences, Agent A was able to retrieve relevant context even for new faults. This allowed it to produce a far more insightful characterization, with the average CSS surging to 4.8. Fig. \ref{fig:qualitative_results} (right panel) provides a direct qualitative illustration of this improvement, where the agent's output is no longer a simple description but a specific, cause-oriented hypothesis that achieves a CSS of 5. This high-quality initial assessment had a direct causal effect on the overall process, reducing the required human interaction by 71\% to an average of just 1.8 dialog turns. These findings provide strong empirical evidence that the Human-in-the-Loop Distillation process successfully creates a virtuous learning cycle, enhancing the AI's perceptual acuity with each human-guided experience.


\section{CONCLUSION AND FUTURE WORK}

We have presented AURA, a novel human-in-the-loop framework that transforms anomaly diagnosis for underwater robots into a collaborative and adaptive process. 
%
%
The key contribution of this work is the \emph{Human-in-the-Loop Distillation} mechanism, which systematically feeds the outcomes of human-AI diagnostic sessions back into the system to refine its knowledge. By splitting responsibilities between a perception-focused agent and a reasoning-focused agent, our architecture ensures that powerful AI tools can be used safely under human supervision, establishing a resilient autonomy paradigm based on continuous learning. This represents a significant step toward achieving robust autonomy in complex robotic operations. 


Our immediate future work will focus on transitioning from laboratory validation to field trials in open marine environments. This will involve deploying AURA in real-world operational scenarios with experienced underwater vehicle engineers, who are not part of the research team, to rigorously evaluate its performance and utility. Open-water deployment will also allow us to quantify the sim-to-real gap of the Stonefish-based digital twin under currents and waves, and to refine its hydrodynamic parameters through online model identification, complementing the adaptive residual estimation already built into the anomaly detector. Furthermore, future work will expand the diversity of tested anomalies and rigorously benchmark AURA's diagnostic accuracy and latency against state-of-the-art model-based and data-driven FDI baselines. 
In parallel, we will investigate automated pre-mission knowledge generation via systematic fault injection and domain randomization in simulation to produce synthetic anomaly cases for VDB pre-population, enabling scalable adaptation to new vehicle morphologies without extensive real-world failure data.
%

\bibliographystyle{IEEEtran}
\bibliography{reference}
\end{document}